\documentclass{article}

\usepackage{spconf,amsmath,graphicx}
\usepackage{xcolor}
\usepackage{multirow}
\usepackage{booktabs}
\usepackage{subcaption}

\usepackage{amsmath}
\usepackage{amsfonts}
\usepackage{graphicx}
\usepackage{xcolor}
\usepackage{soul}
\usepackage{algorithm}
\usepackage{algpseudocode}

\usepackage{tikz}
\usepackage{pgfplots}
\usepackage{pgfplotstable}

\DeclareMathOperator*{\argmax}{\arg\!\max}

\usepackage{array}
\newcolumntype{I}{>{\itshape}c}

\title{Towards matching phones and speech representations}

\name{Gene-Ping Yang, Hao Tang}
\address{Centre for Speech Technology Research, University of Edinburgh}

\copyrightnotice{979-8-3503-0689-7/23/\$31.00~\copyright2023 IEEE}
\begin{document}

\maketitle

\begin{abstract}

Learning phone types from phone instances has been a long-standing problem, while still being open.
In this work, we revisit this problem in the context of self-supervised learning,
and pose it as the problem of matching cluster centroids to phone embeddings.
We study two key properties that enable matching, namely,
whether cluster centroids of self-supervised representations reduce
the variability of phone instances and respect the relationship among phones.
We then use the matching result to produce pseudo-labels and introduce
a new loss function for improving self-supervised representations.
Our experiments show that the matching result captures
the relationship among phones.
Training the new loss function jointly with the regular self-supervised losses, such as APC and CPC,
significantly improves the downstream phone classification.

\end{abstract}

\begin{keywords}
self-supervised learning, acoustic unit discovery, Gromov--Wasserstein distance
\end{keywords}

\section{Introduction}
\label{sec:intro}

Frame representation produced by self-supervised models have shown to work well at distinguishing phone identities at the frame level~\cite{oord2018cpc,chung2019an,yang2022auto,yang21c_interspeech} and at the segment level \cite{chorowski21b_interspeech, bhati2022unsupervised, cuervo2022variable}.
Although phones are thought to be largely separable in the space of speech representation,
the separability is based on \emph{instances}.
Whether it is possible to learn representations of \emph{types} as opposed to instances
remains an open problem.

In this paper, we will focus on phone types, because phonetic properties
are salient even after crude quantization down to 100 or even 50 codes
\cite{hsu2021hubert, wells2022phonetic,Amitay2023}.
The problem of learning phone types has a long history and
shows up under different names, such as acoustic unit discovery \cite{park2008unsup,zero2015,zero2017}
and unsupervised speech recognition \cite{liu18g_interspeech,chen19e_interspeech,liu2022,baevski2021unsupervised,liu2023towards,yeh2018unsupervised}.
It is also highly related to unsupervised phone segmentation \cite{kreuk20_interspeech,zero2015,zero2017}
and lexical discovery~\cite{kamper2017segmental}.

Most, if not all, approaches assume that types arise from clustering of instances.
Approaches based on dynamic time warping for acoustic unit discovery rely on
clustering \cite{park2008unsup}.
Bayesian models for acoustic unit discovery is another form of clustering \cite{lee2012nonparametric}.
Much of the recent progress focuses on adversarial approaches to distinguish
sequences of cluster IDs and phone sequences observed in data sets \cite{liu18g_interspeech,chen19e_interspeech,liu2022,baevski2021unsupervised,liu2023towards}.
In this paper, we follow the same footsteps and study
whether cluster centroids of instances can be
a good representation for types in the context of self-supervised learning.

Even though cluster centroids have been used as types for almost all
approaches to learning phone types,
it is also true that the types discovered by these approaches are never one to one.
There is abundant variability of phones within and across speakers,
and we generally do not know the correct number of clusters to use.
Recent work on analyzing representations of self-supervised predictive coding finds
that phonetic and speaker information are largely represented
in orthogonal subspaces~\cite{liu2023self}.
We will use this as a tool to measure and reduce the variability
of phone instances among speakers.

For the second problem where there is not always a one-to-one mapping
between the centroids and phone types, we take a different route
and assume access to a set of phone embeddings.
We introduce an additional \emph{matching} step to find
the correspondence between the centroids and phone embeddings.
This is reminiscent to the line of work in unsupervised machine
translation~\cite{lample2018word,2018gromov,alaux2018unsupervised}.
We will adopt a similar approach, using optimal transport,
in particular, optimizing the Gromov-Wasserstein
distance~\cite{memoli2011gromov, peyre2016gromov}
to match the centroids and phone embeddings.

We see several applications once we match the centroids and phone embeddings.
One immediate application is to use the matching result
to provide pseudo-labels on unlabeled speech.
The pseudo-labels can be used to further improve the speech representations.
We introduce a loss function predicting pseudo-labels
as self-supervised learning, similar to~\cite{hsu2021hubert,chiu2022self}.

Our experiments show that centroids of self-supervised frame representation
provide a good starting point for learning types.
We will show quantitative and qualitive results for matching centeroids
to phone embeddings.
In addition, by predicting the pseudo-labels produced by the
matching result gives a sizable improvement in
downstream phone classification.

\section{Proposed Approach}

\begin{figure}[t]
    \centering
    \includegraphics[width=0.8\linewidth]{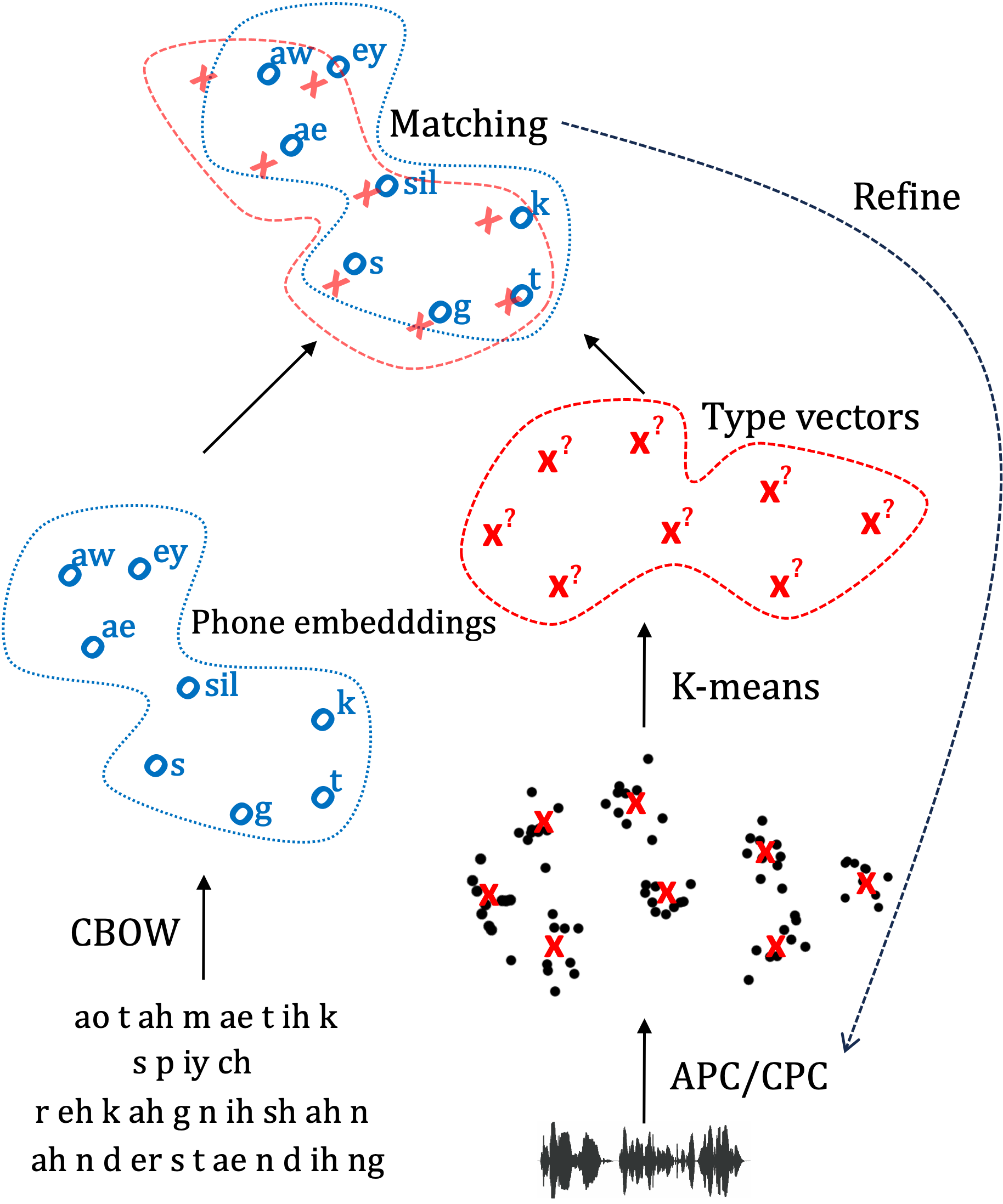}
    \caption{An overview of our proposed approach. Phone embeddings are learned from phone sequences with CBOW.
    Type vectors are derived from clustering self-supervised frame representations.
    Matching assigns phone labels to each centroid and can be used to provide pseudo-labels for unlabeled speech.
    Based on the matching result, an additional objective is introduced for self-supervised learning that predicts the pseudo-labels.}
    \label{fig:overview}
\end{figure}

An overview of our proposed approach is illustrated in Figure \ref{fig:overview}.
The goal is to match type vectors derived from a self-supervised model to phone embeddings
(without using paired transcriptions).
Vectors produced by self-supervised models are contextualized, meaning that different phone
instances would have different sequences of vectors.
Since vectors produced by self-supervised models tend to be sufficiently clustered by phones \cite{hsu2021hubert},
we make the first assumption that a type can be represented by the mean of its instances.
As a result, we define \textbf{type vectors} to be centroids after clustering
frame vectors produced by self-supervised models.

We further make a second assumption that the derived type vectors respect the relationship among phones,
for example, similar pronouncing phones are closer in space.
If we have access to phone embeddings that also respect the relationship of phones,
the two sets of vectors have the potential to be matched.
In this work, we use phone embeddings from a continuous bag-of-word (CBOW) model \cite{tomas2013,tomas2013-2}
trained on phone sequences extracted from texts and a pronunciation dictionary.

To match the two sets of vectors, we use optimal transport, in particular, optimizing the
Gromov-Wasserstein distance \cite{memoli2011gromov, peyre2016gromov}.
Finally, with the goal of improving speech representations,
we infer the matched phone labels for every frame,
and treat them as pseudo-labels for self-supervised learning.

Our approach is unsupervised, in the sense that it does not require manual transcriptions paired with speech.
The dependency on texts and a pronunciation dictionary may seem stringent.
However, as we will show in the experiments, we observe improvements even
when the matching result is only approximate.
This suggests that the requirement on texts and a pronunciation dictionary
can be relaxed, and we leave it as future work.

\section{Assumptions and Validation}

There are two main assumptions in our approach: one is that the mean of instances would be representative of the instances and the other is that type vectors respect the relationship of phones.
In the first assumption, whether the mean is representative of the instances depends on the variability of frame representations.
In the second assumption, whether the relationship of phones are respected depends on how they are organized in space and their nearest neighbors.
We will study both assumptions in this section.

\subsection{Variability of Phone Instances}

To study the variability of phone instances, inspired by \cite{liu2023self}, we collect all frame vectors produced by a self-supervised model of the same phone label into a matrix and compute PCA on it.
Specifically, suppose all frames produced by a self-supervised model are forced aligned to phones.
We compute PCA on the matrix $M^{p}$ whose column vectors are frames labeled as phone $p$ in a data set.
In other words, the set of eigenvectors we get spans the subspace of a particular phone
and captures the variance of the instances of that particular phone.

Similarly, we define the speaker subspace as the span of speaker vectors, each of which is an average of frame vectors that have the same speaker label.
We compute PCA on $M^{\text{spk}} = [m_{s_1}, \dots, m_{s_J}]$ for a set of $J$ speakers $\{s_1, \dots, s_J\}$ where $m_s = \text{avg}(\{h: \text{the frame $h$ is from speaker $s$}\})$.
We compute PCA on the speaker vectors as well to obtain eigenvectors for the speaker subspace.
Given two sets of eigenvectors (one from phone instances and one from speakers),
we can study how correlated they are.

\begin{figure}
    \centering
    \begin{subfigure}[b]{0.23\textwidth}
        \centering
        \includegraphics[width=\textwidth]{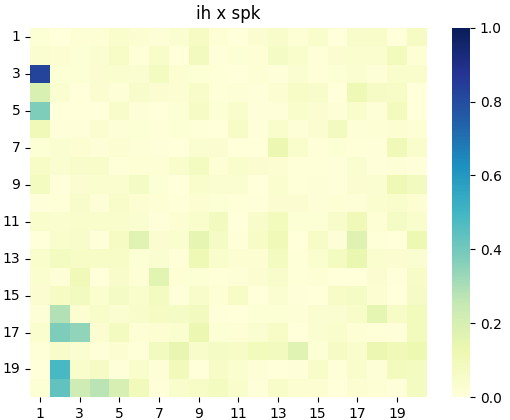}
        \caption{before collapsing}    
        \label{fig:mean and std of net14}
    \end{subfigure}
    \begin{subfigure}[b]{0.23\textwidth}   
        \centering 
        \includegraphics[width=\textwidth]{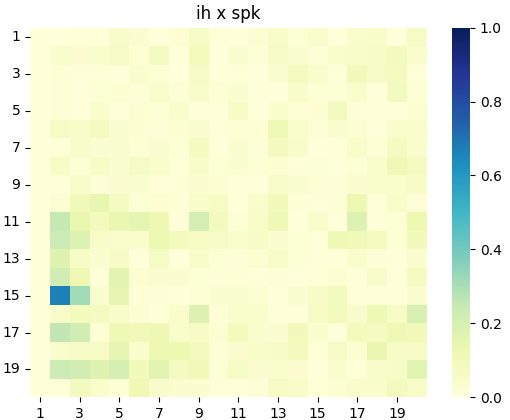}
        \caption{after collapsing}    
        \label{fig:mean and std of net34}
    \end{subfigure}
    \caption{The absolute of dot products among the top 20 eigenvectors from the APC speaker subspace (x-axis) and those from the frame representation of the phone [ih] (y-axis). (a) The absolute of dot products before collapsing the first speaker direction. (b) The same measures after collapsing the first speaker direction.}
    \label{fig:phone_space}
\end{figure}

We conduct our analysis on the 100-hour subset of Librispeech with 3-layer LSTMs trained with APC
\cite{yang2022auto} and CPC~\cite{oord2018cpc} (more details in the experimental section).
In Figure~\ref{fig:phone_space}, we focus on APC and use the phone [ih] as an example
and show the absolute of dot products of the two sets of eigenvectors.
In Figure \ref{fig:phone_space} (a),
the first speaker direction (ranked third) contributes significantly
to the variability of the phone instances.\footnote{The first speaker direction likely indicates the average fundamental freqeuencies, separating males and females \cite{liu2023self}.}
We adopt the idea in \cite{liu2023self}, projecting a frame vector $h$ to a subspace
orthogonal to the first speaker direction $v$ (called collapsing in \cite{liu2023self})
\begin{align}
    h_c = h - (h^\top v) v,
\end{align}
where $h_c$ is the collapsed frame vector.
The resulting absolute of dot products is shown in Figure \ref{fig:phone_space} (b).
Collapsing effectively reduces the variability along the speaker directions.
We could continue collapsing but the second speaker direction (ranked 15th) contributes significantly less than the first.

\begin{figure}
    \centering
    \includegraphics[width=0.6\linewidth]{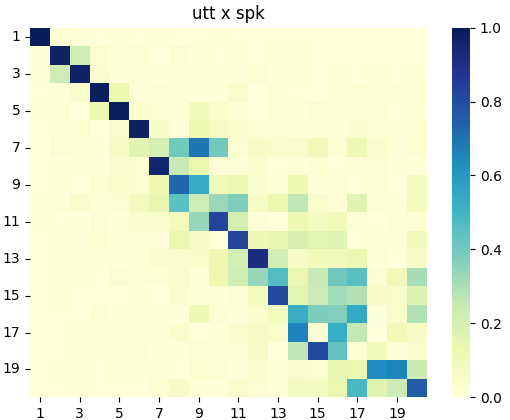}
    \caption{The absolute of dot products comparing the top 20 principle directions of the APC speaker subspace (x-axis) and those from the utterance subspace (y-axis).}
    \label{fig:spkxutt}
\end{figure}

The collapsing approach relies on speaker labels that are not always at hand.
We discover an alternative approach to find the speaker direction without using speaker labels.
Instead of averaging frame vectors based on speakers, we average frame vectors based on utterances and compute PCA on it.
The absolute of dot products of the two approaches are shown in Figure \ref{fig:spkxutt},
and the eigenvectors are surprisingly well aligned for the first few directions.

The findings hold for most phones and for both APC and CPC.
Later in the experiments, we will approximate speaker
directions with utterances and collapse the first speaker dimension before clustering.
Controlling variability of instances is deemed essential in unsupervised
automatic speech recognition.
For example, \cite{baevski2021unsupervised} relies on silence removal,
dimensionality reduction with PCA after k-means, and mean pooling.

\subsection{Nearest Neighbors of Phone Types}

To study whether the type vectors respect the relationship of phones,
we use forced alignments (only for analysis) and compute the means of frame vectors that have the same phone label.
Table \ref{tab:phone_nn}, based on APC, shows the nearest neighbors of the means and the phone embeddings learned
from CBOW.
We observe that the vectors in the two spaces share similar nearest neighbors.
In addition, vowels share nearest neighbors with vowels, while consonants
share nearest neighbors with consonants.
Manners and places of articulation for consonants also form groups.
This provides us evidence that there is enough signal to match the two
spaces.
It also suggests that an accurate matching is not required (if possible at all).
An approximate match would provide training signals to distinguish vowels and consonants
or even the manners and places.

\begin{table}
    \caption{Top 3 nearest neighbors of phone types from APC and CBOW.
    Phone types from APC are the averages of frame representation
    (of the same phone label) based on forced alignments.
    Nearest neighbors that appear in both sets are underlined.}
    \label{tab:phone_nn}
    \centering
    \scalebox{0.8}{
    \begin{tabular}{c|cc|c|cc}
        \toprule
        Phone & APC & CBOW & Phone & APC & CBOW \\
        \midrule
        ae    & eh \underline{aw} ay & ih ow \underline{aw} & 
        k  & g, \underline{t}, p & m, \underline{t}, f \\
        ah    & spn, \underline{ih}, uh & er, \underline{ih}, ow        & 
        m  & n, spn, b & t, s, l \\
        aw    & aa, \underline{ae}, eh & \underline{ae}, ay, ih         & 
        p  & b, k, \underline{t} & s, \underline{t}, m \\
        er    & r, \underline{spn}, \underline{ah} & \underline{ah}, \underline{spn}, ih      & t  & \underline{d}, spn, jh & \underline{d}, m, z\\
        ey    & iy, ih, \underline{eh} & \underline{eh}, aa, ah         & 
        th & f, dh, \underline{t} & \underline{t}, z, ch \\
        ih    & \underline{ah}, spn, eh & \underline{ah}, ow, ae        & 
        z  & \underline{s}, spn, \underline{t} & d, \underline{t}, \underline{s} \\
        ch    & jh, \underline{t}, sh & \underline{t}, m, z             & 
        s  & \underline{z}, spn, t & m, \underline{z}, b \\
        d     & \underline{t}, spn, n & \underline{t}, z, m             & 
        sh & \underline{ch}, zh, jh & t, s, \underline{ch}\\
        g     & \underline{k}, b, d & \underline{k}, f, b               & 
        v  & \underline{b}, spn, f & \underline{b}, m, n \\
        jh    & ch, \underline{t}, d & \underline{t}, v, b              & 
        zh & sh, \underline{jh}, ch & \underline{jh}, t, f \\
        \bottomrule
    \end{tabular}
    }
\end{table}

\section{Matching}

There are many algorithms that aim to match two sets of vectors.
Based on the analyses in the previous section,
we choose to optimize the Gromov--Wasserstein distance \cite{memoli2011gromov,peyre2016gromov}
as it respects distances among vectors and does not assume a linear transformation between
the two spaces.
Suppose the set of vectors $\{c_1, \dots, c_n\}$ are the centroids after clustering
the frame vectors produced by a self-supervised model,
and the set of vectors $\{y_1, \dots, y_m\}$ are the phone embeddings learned
from CBOW.
Note that $n$ and $m$ are not required to be the same.
We compute two distances matrices $S$ and $S'$ on the two spaces,
where $S_{ij} = \| c_i - c_j \|_2^2$ and $S'_{ij} = \| y_i - y_j \|_2^2$.
The goal is to find $\Gamma$ where $\Gamma_{ij}$ indicates
the probability of matching $c_i$ to $y_j$.
When $c_i$ is matched to $y_j$ and $c_k$ is matched to $y_l$, we incur a cost $(S_{ik} - S'_{j\ell})^2$
for not respecting the distances.
Based on these definitions, the Gromov-Wasserstein distance is defined as
\begin{align*}
    \min_\Gamma & \quad \sum_{ij} \sum_{k\ell} (S_{ik} - S'_{j\ell})^2 \Gamma_{ij} \Gamma_{k\ell} - \epsilon H(\Gamma) \\
    \text{s.t.} & \quad \Gamma \mathbf{1} = p, \quad \Gamma^\top \mathbf{1} = q, \quad \Gamma \geq 0
\end{align*}
where $\mathbf{1}$ is an all-one vector, $H(\Gamma)$ is the sum of entropy for all rows in $\Gamma$, and $\epsilon$ is the entropy regularization parameter.
The objective additionally has two unigram distribution vectors $p$ and $q$.
In our case, $p$ is the unigram distribution of centroids, and $q$ is the unigram
distribution of phones in the texts for training CBOW.
Both can be computed without relying on manual transcriptions or forced alignments.

To optimize the Gromov--Wasserstein distance, following~\cite{peyre2016gromov}, we iteratively compute a loss matrix $L$ and the matching matrix $\Gamma$ as follows.
\begin{align}
    L & \gets S^2 p \mathbf{1}^\top + \mathbf{1} q^\top {S'^2}^\top- 2 S \Gamma S'^\top \\
    a & \leftarrow p \oslash Kb, \quad b \leftarrow q \oslash K^\top a \\
    \Gamma & \gets \text{diag}(a) K \text{diag}(b)
\end{align}
where $K=e^{-L / \epsilon}$.
We technically do not need to learn a linear transformation between the two spaces
(if the two spaces can be matched at all with a linear transformation).
If this is necessary, we can use Procrustes analysis to find the linear transformation between the two spaces.
Specifically, if the SVD of $C \Gamma Y^\top$ is $U \Sigma V^\top$,
the linear transformation between the two spaces is $UV^\top$.

\section{Pseudo-labels from Matching}

Once matching is done, every centroid is matched to a phone embedding.
We can label each frame vector based on which centroid it belongs and
what phone embedding it gets matched.
This process creates sequences of pseudo-labels for unlabeled speech
without relying on manual transcriptions.
With the goal of improving speech representations,
we can create a new self-supervised loss function, predicting the pseudo-labels
discovered with this process.

More formally, we assign each frame at $t$ a pseudo-label
\begin{align}
s_t = \argmax_i \Gamma_{{z_t}, i}
\end{align}
where $z_t$ is the centroid that the frame belongs.
In other words, the pseudo-label is the matched phone given the centroid that the frame belongs.
Predicting pseudo-labels is a common approach to self-supervised learning.
For example, wav2vec 2.0 uses the quantized output of the convolution layers as pseudo-labels \cite{BZMA2020},
while HuBERT uses quantized MFCCs or one of its own layers as pseudo-labels \cite{hsu2021hubert}.
There is even evidence that randomly quantizing log Mel features (also known as BEST-RQ in  \cite{chiu2022self}) could serve as pseudo-labels for self-supervised learning.

When predicting the pseudo-labels, we have two options: one is to use cross entropy loss and the other is to get close to the target embedding with the $\ell_2$ loss.
For cross entropy loss, we add the loss
\begin{align}
\sum_{t=1}^{T-k} \mathbf{1}_{s_{t+k}}^\top \log \text{softmax}(W h_t),
\end{align}
to measure the prediction quality,
where $\mathbf{1}_i$ is a one-hot vector with the $i$-th dimension set to 1, $h_t$ is the frame vector produced by a self-supervised model at time $t$, $W$ is for linear prediction,
and $s_{t+k}$ is the pseudo-label.
As an aside, if $s_{t+k}$ is the cluster ID of clustering log Mel features with a random codebook, this approach becomes a special case of BEST-RQ~\cite{chiu2022self}.

In the second case, more formally, recall that $y_i$ is the $i$-th phone embedding from CBOW.
To predict the target embedding, we will add
\begin{align}
\sum_{t=1}^{T-k}\| y_{s_{t+k}} - W h_t\|^2_2
\end{align}
to the training objective
where $W$ is for linear regression,
and $y_{s_{t+k}}$ is the embedding looked up with the pseudo-label $s_{t+k}$.
Note that we have applied a time shift of $k$ on the targets to be consistent with APC and CPC.

\section{Experiments}
\label{sec:exp}

Following the protocol of \cite{yang2022auto},
we evaluate our proposed approach by pre-training on the 360-hour subset of LibriSpeech
and performing phone classification on Wall Stree Journal (WSJ).
The pre-trained models are frozen after pre-training, as we are interested in the quality
of speech representations after pre-training.
Ten percent of the training set (si284) in WSJ is used as a development set for choosing the best phone classifier.
We report the phone error rates of the selected probing models on dev93 and eval92. 
A phone set with 41 phones are used in our study, including one label (sil) for silence and one (spn) for spoken noise.
To compare against prior work, we use 3-layer uni-directional LSTMs with a hidden size of 512 dimension in all experiments.
To test the generalization across self-supervised losses, we compare both APC \cite{yang2022auto} and CPC \cite{oord2018cpc}.
We choose a time shift of 5 for both APC and CPC, and train them with a batch size of 32 and a step size of $10^{-3}$ using the Adam optimizer.
For CPC, following \cite{chung2019an}, we do not use convolution layers and directly contrast on log Mel features.
A set of 100 negative samples are uniformly drawn from the log Mel frames of the same utterance.
All models are trained with 15 epochs for pre-training, and another 15 epochs for phone classification.

For CBOW on phone sequences, we use the texts on the 100-hour subset of Librispeech
and CMUdict version 0.7.
We train CBOW with a window size of 5, a step size of 0.005, a batch per utterance,
and gradient clipping to norm 5.\footnote{Incidentally, there is little prior work on CBOW for learning phone embeddings. There was a blog post on this by Gabriel Synnaeve written in 2014, and an unpublished manuscript by Michael Hammond in 2020.}
Given the relatively small number of classes, we consider all classes and do not use negative samples.

\subsection{Matching with Gromov-Wasserstein Distance}

Before matching, we first evaluate the quality of clustering frame vectors produced by self-supervised models.
Following~\cite{hsu2021hubert, wells2022phonetic}, we analyze the
phone purity (PP), cluster purity~(CP) and phone error rates (PER) with forced alignments.
Note that phone purity first computes a purity for each individual cluster and averages them,
while frame phone error rates is the total number of errors summed over all clusters divided by the total number number of frames.

We run Lloyd's algorithm for k-means clustering with 50 clusters on the 100-hour subset of LibriSpeech for 20 epochs. 
We choose the second LSTM layer for APC and CPC, as well as the ninth transformer layer for HuBERT,
because the phonetic information is most prominent in these layers \cite{yang2022auto, hsu2021hubert}.
For APC and CPC, we also include the ones after collapsing the first speaker direction (denoted APC$_c$ and CPC$_c$).
We include random projection on log Mel features \cite{chiu2022self},
initializing the projection matrix using the Glorot initialization \cite{glorot2010under} and the codebook with a unit Gaussian.

Results are shown in Table \ref{tab:kmeans}.
Surprisingly, similar performance has been observed among APC, CPC and HuBERT,
despite that HuBERT is much larger in size and trained with more data.
There is a drastic difference in phone purity and frame PER for random projection.
This is due to large imbalance in cluster sizes.
We will leave HuBERT and random projection for future work and focus on APC and CPC
for the rest of the paper.

\begin{table}
    \centering
    \caption{Analyses on clusters and matching with forced alignments.
    The evaluation metrics are phone purity (PP), cluster purity (CP), and frame and type phone error rates (PER).
    We compare APC, CPC, and their counterpart APC$_c$ and CPC$_c$ after collapsing the first speaker direction.
    HuBERT and random projection are also listed.}
    \begin{tabular}{ccccc}
        \toprule
        frame & \multirow{2}{*}{PP ($\uparrow$)} & \multirow{2}{*}{CP ($\uparrow$)} & frame & type \\
        vectors & & & PER ($\downarrow$) & PER ($\downarrow$) \\
        \midrule
        APC &  50.1 & 37.3 & 50.1 & 93.1 \\ 
        APC$_c$ & 47.5 & 36.6 & 50.1 & 76.8\\
        CPC &  51.1 & 37.7 & 48.9 & 73.0 \\ 
        CPC$_c$ & 50.6 & 38.2 & 47.8 & 72.6\\
        HuBERT & 50.3 & 37.8 & 48.3 & - \\
        Rand Proj & 22.1 & 29.5 & 73.0 & - \\
        \bottomrule
    \end{tabular}
    \label{tab:kmeans}
\end{table}

Next, we evaluate the quality of matching centroids to phone embeddings with Gromov--Wasserstein distance.
We compute type phone error rates, i.e., how often an incorrect phone label is assigned to a centroid.
The ground truth label of a centroid is determined
with the majority vote of frames based on forced alignments.
Prior to computing the distance matrices, we center and normalize the vectors to unit norm, following \cite{2018gromov,alaux2018unsupervised}.
We use entropy regularization of 0.0005 for APC and 0.01 for CPC.
The iterative algorithm is run for 1,000 iterations.
The unigram distribution for both the centroids and the phones are
computed on the respective data sets that they are trained on.
Results are shown in the last column of Table \ref{tab:kmeans}.
Collapsing the first speaker direction improves both frame and type PERs.

\begin{figure}[!htb]
    \centering
    \begin{subfigure}[b]{0.85\linewidth}
        \centering
        \includegraphics[width=\textwidth]{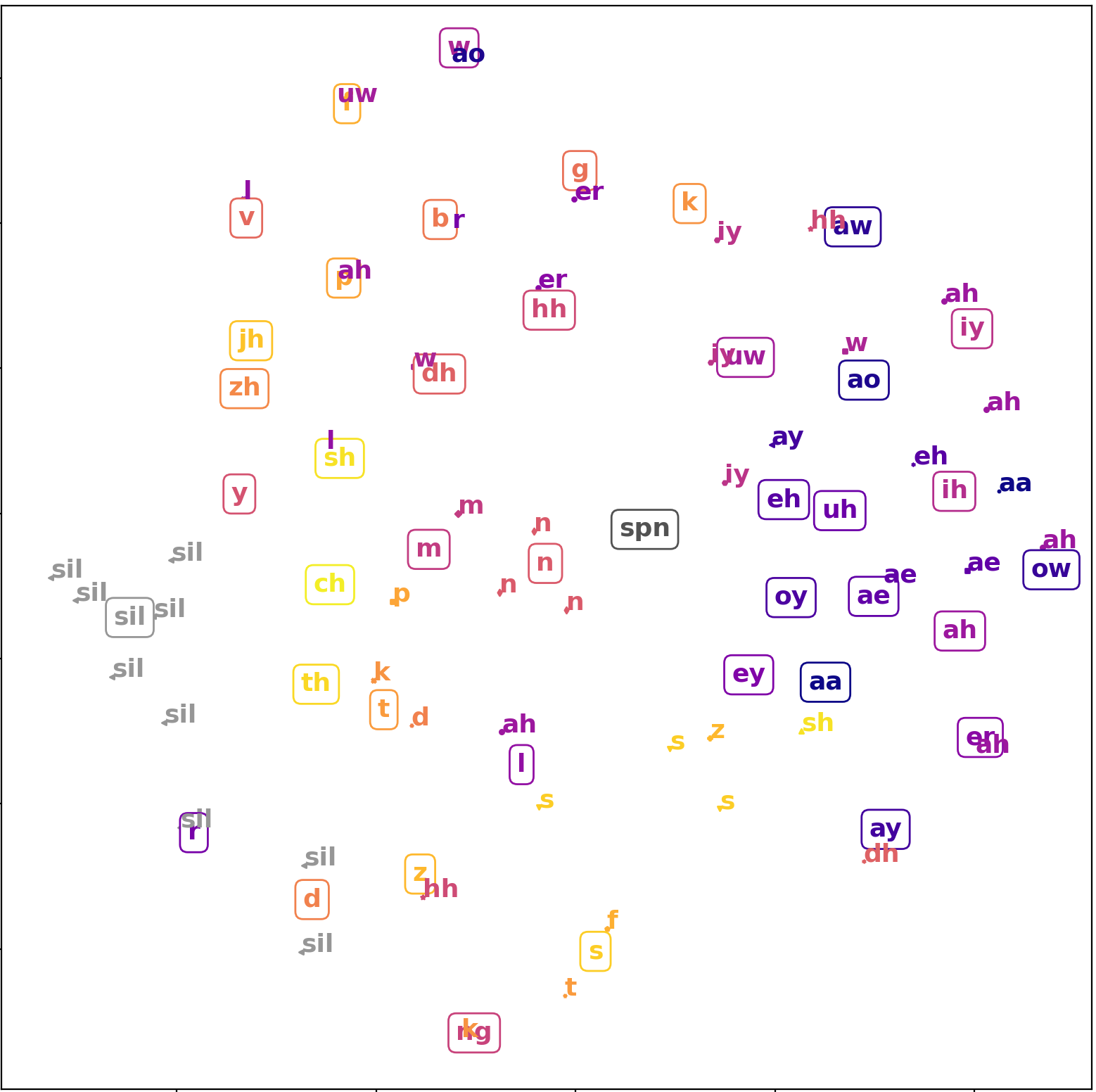}
    \end{subfigure}
    \caption{A t-SNE plot of the matching result.
    The type vectors are projected onto the space of phone embeddings using a linear transformation learned with Procrustes analysis.
    Phone embeddings are boxed, while the centroids are not.
    The color uses the first phone direction in \cite{liu2023self} that correlates
    with sonority.}
\label{fig:tsne_apc}
\end{figure}

To evaluate the matching result qualitatively, we use Procrustes analysis to learn the linear transformation
of the two spaces and plot the two spaces together with tSNE.
The visualization for APC is shown in Figure \ref{fig:tsne_apc}.
Though the type PERs are high, the matching is qualitatively successful in grouping
vowels, consonants, and silence.
Within consonants, fricatives, stops, and nasals also form groups.

\subsection{Self-Supervised Learning with Pseudo-Labels}

Though matching between centroids and phone embeddings produces high type PERs,
there are relationship among phones recovered from matching is still valuable
for speech representation learning.
We produce pseudo-labels for the 360-hour subset of LibriSpeech
by looking up the centroid each frame belongs to and the phone embedding
it is matched.
We introduce a new self-supervised loss function by predicting the target
embeddings of the pseudo-labels.

Note that if we obtain a perfectly accurate match between the two spaces,
the task of predicting pseudo-labels falls back to supervised phone
classification.
Based on this observation, we first conduct a topline of jointly
training with a self-supervised loss function and a supervised loss
function \cite{wang2021unispeech,bai2022joint}.
We further conduct a set of experiments, adding corruption to phone
labels to simulate joint training with matching.
Note that when the percentage of corruption is 100\%,
we randomly sample a phone embedding to predict, so it
is different from BEST-RQ \cite{chiu2022self} that predicts codeword vectors from
a random codebook.

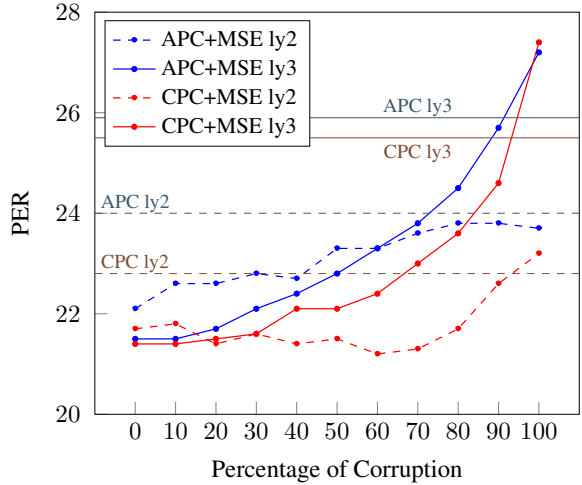
\begin{figure}
    \centering
    \pgfplotstableread{ 
         0  22.1  21.5  21.7  21.4
        10  22.6  21.5  21.8  21.4
        20  22.6  21.7  21.4  21.5
        30  22.8  22.1  21.6  21.6
        40  22.7  22.4  21.4  22.1
        50  23.3  22.8  21.5  22.1
        60  23.3  23.3  21.2  22.4
        70  23.6  23.8  21.3  23.0
        80  23.8  24.5  21.7  23.6
        90  23.8  25.7  22.6  24.6
       100  23.7  27.2  23.2  27.4
    }\dataset
    \resizebox{0.9\linewidth}{!}{%
    \begin{tikzpicture}
        \begin{axis}[
            xlabel={Percentage of Corruption},
            ylabel={PER},
            ylabel style={align=center, yshift=-5pt},
            ymax=28,
            ymin=20,
            xtick={0, 10, 20, ..., 100},
            xtick pos=left,
            ytick pos=left,
            grid style=dashed,
            yticklabel style={xshift=-0.5ex},
            tickwidth=5pt,
            minor x tick style = {opacity=0},
            legend style={at={(0.02,0.98)}, anchor=north west, font=\small},
            ]
            \addplot[color=blue, line width=0.5pt, mark=*, mark size=1pt, dashed] table[y index=1] \dataset;
            \addlegendentry{APC+MSE ly2}
            \addplot[color=blue, line width=0.5pt, mark=*, mark size=1pt] table[y index=2] \dataset;
            \addlegendentry{APC+MSE ly3}
            \addplot[color=red, line width=0.5pt, mark=*, mark size=1pt, dashed] table[y index=3] \dataset;
            \addlegendentry{CPC+MSE ly2}
            \addplot[color=red, line width=0.5pt, mark=*, mark size=1pt] table[y index=4] \dataset;
            \addlegendentry{CPC+MSE ly3}
            \addplot [color={rgb:red,36;green,54;blue,66}, line legend, sharp plot, update limits=false, dashed] coordinates {(-10, 24.0) (110, 24.0)}
            node at (0, 420) {\footnotesize APC ly2};
            \addplot [color={rgb:red,36;green,54;blue,66}, line legend, sharp plot,update limits=false] coordinates { (-10, 25.9) (110, 25.9) }
            node at (70,610) {\footnotesize APC ly3};
            \addplot [color={rgb:red,100;green,60;blue,40}, line legend, sharp plot,update limits=false, dashed] coordinates { (-10, 22.8) (110, 22.8) }
            node at (0, 300) {\footnotesize CPC ly2};
            \addplot [color={rgb:red,100;green,60;blue,40}, line legend, sharp plot,update limits=false] coordinates { (-10, 25.5) (110, 25.5) }
            node at (70,520) {\footnotesize CPC ly3};
        \end{axis}
    \end{tikzpicture}
    }
    \caption{Phone error rates (PER) for evaluating joint self-supervised and supervised pre-training under different percentages of corruption to forced alignments.
    Both layer 2 (ly2) and layer 3 (ly3) of LSTMs are reported.
    We measure the prediction of phone embeddings with $\ell_2$ loss, hence the label MSE.}
    \label{fig:perturb}
\end{figure}

Results of adding corruption to phone labels during
joint self-supervised and supervised training are shown in Figure \ref{fig:perturb}.
In particular, since we get about 77\% type errors in Table~\ref{tab:kmeans} for APC,
we are interested in the case where the pseudo-labels have high type errors.
We observe improvements in phone classification even at a high percentage of corruption.
This suggests that even when the matching result has a type PER of 77\%,
we can still expect improvements.

\begin{table}
  \caption{Phone error rates (PER) for the downstream phone classification on WSJ.
  The additional loss function is annotated with MSE for the $\ell_2$ against the target embeddings
  and CE for the cross entropy loss against the codes.
  The settings for forced alignments with 77\% corruption (FA @ 77\%) are also listed.}
  \label{tab:final}
  \centering
  \scalebox{0.8}{
  \begin{tabular}{lccccc}
    \toprule
    \multirow{2}{*}{objective} & \multirow{2}{*}{pseudo-labels} & \multicolumn{2}{c}{layer 2} & \multicolumn{2}{c}{layer 3} \\
    & & dev93 & eval92 & dev93 & eval92 \\
    \midrule
    APC & - & 22.0 & 22.0 & 23.8 & 24.2 \\
    APC + MSE & Matching & \textbf{20.5} & \textbf{19.9} & 22.5 & 22.2 \\
    APC + MSE & Rand Proj & 22.5 & 22.3 & 26.8 & 28.5 \\
    APC + CE & Rand Proj & 21.3 & 21.1 & 24.5 & 24.8 \\
    \midrule
    CPC & - & 20.7 & 20.3 & 23.2 & 23.4 \\
    CPC + MSE & Matching & \textbf{20.0} & \textbf{19.6} & 21.9 & 21.8 \\
    CPC + MSE & Rand Proj & 22.5 & 22.7 & 28.0 & 29.5 \\
    CPC + CE & Rand Proj & 20.8 & 20.4 & 24.2 & 24.8 \\
    \midrule
    APC + MSE & FA @ 77\% & 21.4 & 21.2 & 22.3 & 22.2 \\
    CPC + MSE & FA @ 77\% & 19.5 & 19.2 & 21.2 & 20.9 \\    
    \bottomrule
  \end{tabular}
  }
\end{table}

Results of predicting pseudo-labels derived from matching are shown in Table \ref{tab:final}.
We observe a 1.5\% and 2.1\% absolute improvement over regular APC on dev93 and eval92.
The improvements are observed for both layer 2 and layer 3.
The results are also similar when training with CPC.
We include random projection on log Mel features as another set of baseline.
We follow \cite{chiu2022self}, quantizing log Mel features
with a random codebook sampled from unit Gaussian.
We also include a variant where we use $\ell_2$ (MSE) as the loss and the codeword vectors as targets,
to compare with the matching result using the same loss function. 
Our proposed approach is better than random projection.
Finally, we list the result of corrupting 77\% of the phone labels based on forced alignments.
Jointly predicting pseudo-labels and training with APC actually
performs better, while the CPC counterpart is not far behind.

\section{Conclusion}

In this paper, we have studied the space of self-supervised representations and presented a novel approach to match frame representations to phone embeddings without paired manual transcriptions. 
While the type error rates are high after matching, qualitative analysis shows promising results.
Using the matching result to generate pseudo-labels as prediction targets,
we observe a sizeable improvement on the downstream phone classification. 
The approach can be further refined by iterating between matching and self-supervised learning.
Our positive result also suggests that it might be possible to jointly
perform matching while optimizing a self-supervised loss function.

\bibliographystyle{IEEEbib}
\bibliography{refs}

\end{document}